\documentclass[runningheads]{llncs}

\usepackage{eccv}

\usepackage{eccvabbrv}
\usepackage{graphicx}
\usepackage{booktabs}
\usepackage[accsupp]{axessibility}

\usepackage{hyperref}

\usepackage{orcidlink}

\begin{document}

\title{CalibFree: Self-Supervised View Feature Separation for Calibration-Free Multi-Camera Multi-Object Tracking}

\titlerunning{CalibFree}

\author{
Ruiqi Xian\inst{1} \and
Deep Patel\inst{2} \and
Iain Melvin\inst{2} \and
Sanjoy Kundu\inst{3} \and
Martin Renqiang Min\inst{2} \and
Dinesh Manocha\inst{1}
}

\authorrunning{R. Xian et al.}

\institute{
University of Maryland, College Park, MD, USA \and
NEC Laboratories America, Princeton, NJ, USA \and
University of North Carolina, Greenboro, NC, USA
}

\maketitle

\begin{abstract}
Multi-camera multi-object tracking (MCMOT) faces significant challenges in maintaining consistent object identities across varying camera perspectives, particularly when precise calibration and extensive annotations are required. In this paper, we present CalibFree, a self-supervised representation learning framework that does not need any calibration or manual labeling for the MCMOT task. By promoting feature separation between view-agnostic and view-specific representations through single-view distillation and cross-view reconstruction, our method adapts to complex, dynamic scenarios with minimal overhead. Experiments on the MMP-MvMHAT dataset show a 3\% improvement in overall accuracy and a 7.5\% increase in the average F1 score over state-of-the-art approaches, confirming the effectiveness of our calibration-free design. Moreover, on the more diverse MvMHAT dataset, our approach demonstrates superior over-time tracking and strong cross-view performance, highlighting its adaptability to a wide range of camera configurations. Code will be publicly available upon acceptance.
\end{abstract}

\section{Introduction}
\label{sec:intro}

\begin{figure}[tp]
    \centering
    \includegraphics[width=0.85\columnwidth]{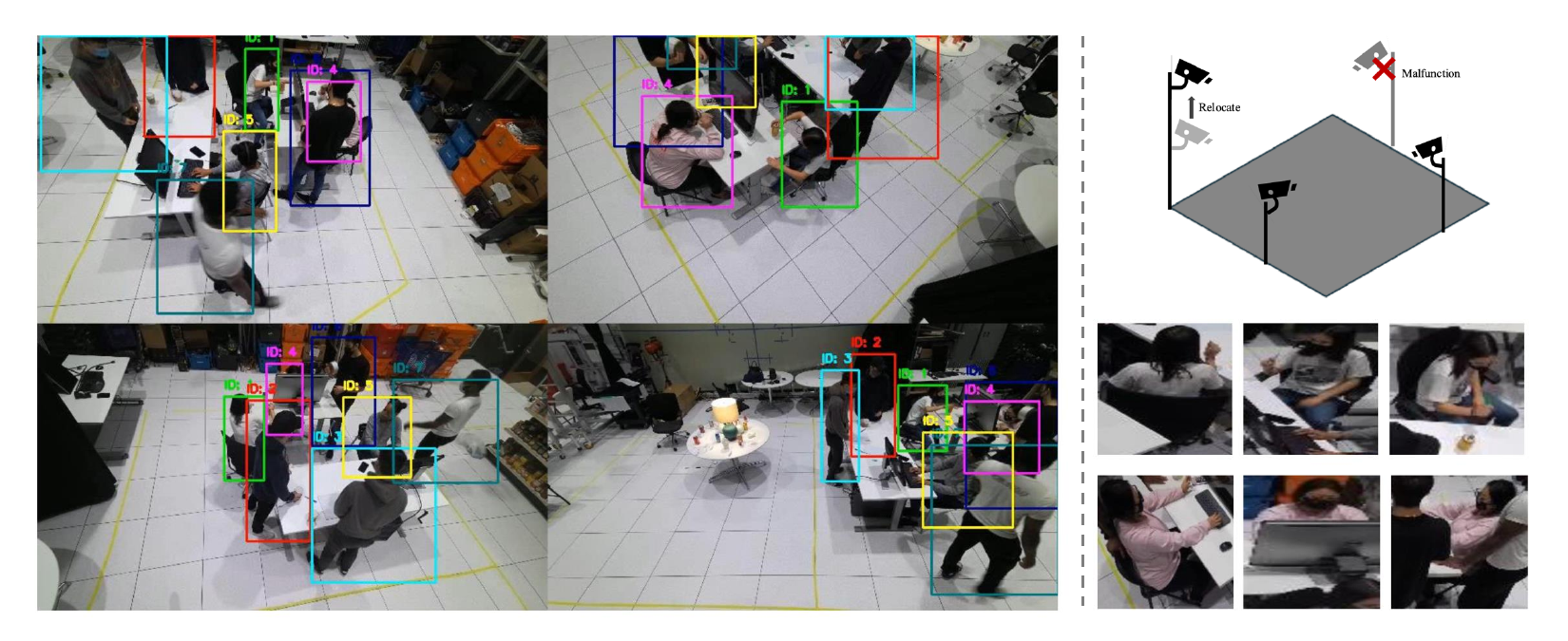}
    \vspace{-10pt}
    \caption{\textbf{Multi-Camera Multi-Object Tracking (MCMOT) setup.} \textit{Left:} Multi-view scenes with individuals tracked across overlapping camera views, each assigned a unique color-coded bounding box. \textit{Top right:} Flexible camera configurations illustrating variable camera numbers and placements across scenarios, or due to factors like malfunction or relocation. \textit{Bottom right:} Examples of appearance variations for individuals across different viewpoints, highlighting the challenge of maintaining consistent identity association in multi-view tracking.
}
    \label{fig:vis_data} 
    \vspace{-15pt}
\end{figure}

Multi-Object Tracking (MOT) has seen rapid progress in recent years~\cite{cao2023observation,wojke2017simple,cai2022memot,meinhardt2022trackformer,zeng2022motr,zhang2023motrv2}. With the increasing deployment of camera networks in surveillance, transportation, and smart-city environments, Multi-Camera Multi-Object Tracking (MCMOT) has become essential for enabling long-range identity preservation and reducing occlusions and blind spots~\cite{gilbert2006tracking,he2020multi,you2020real,cheng2023rest,zhang2022mutr3d,gu2023vip3d}. By integrating complementary viewpoints, MCMOT provides richer scene understanding and more reliable tracking, as illustrated in Fig.~\ref{fig:vis_data}.

However, achieving robust MCMOT in real-world deployments remains a challenge. Cross-camera observations often differ substantially due to viewpoint changes, occlusions, and scene-specific visual biases, making consistent identity representation nontrivial. Many existing approaches rely on calibrated camera setups or scene-specific geometric priors to establish cross-view correspondences~\cite{ristani2016performance,maksai2017non,tesfaye2019multi,he2020multi,you2020real,cheng2023rest}. However, camera networks are far from static: cameras drift over time, undergo maintenance, or experience subtle repositioning. Such changes quickly invalidate calibration and degrade performance, rendering continual reannotation or recalibration costly and impractical at scale.

Although recent efforts have investigated calibration-free MCMOT through geometric reasoning or graph-based formulations~\cite{quach2021dyglip,nguyen2022lmgp}, these methods still implicitly assume relatively stable camera networks. Geometry-driven approaches remain sensitive to camera pose changes, while graph-based models---trained on fixed viewpoints and supervised identity features---can struggle under large view or scene shifts. In dynamic environments where cameras frequently adjust or evolve, these assumptions are difficult to maintain.

These limitations motivate a fundamentally different direction. We argue that a purely RGB-driven, self-supervised representation is essential for achieving truly calibration-free MCMOT. In multi-camera systems with partially overlapping fields of view---common in surveillance and smart-city deployments---visual co-occurrences naturally provide the supervisory signal needed to learn cross-view alignment without labels or geometric constraints. By leveraging these overlaps, our framework learns directly from raw visual co-occurrence rather than relying on fixed geometric relationships or supervised identity labels, making it inherently resilient to camera drift, viewpoint changes, and partial network failures. Moreover, unlike prior calibration-free formulations that incorporate structural priors or implicit geometry, a self-supervised approach naturally adapts to new scenes and camera configurations without requiring annotations or camera parameters. In evolving multi-camera deployments, self-supervision is therefore not merely an alternative---it is a necessary foundation for scalable, maintenance-free tracking.

\paragraph{Main Contribution.}
Motivated by these challenges, we propose a fully self-supervised representation learning framework for MCMOT that operates without any camera calibration, geometric constraints, or manual annotations. Our key idea is to encourage representation specialization for cross-view consistent identity cues and camera-dependent appearance factors. Concretely, we learn view-agnostic and view-specific representations with distinct roles for cross-camera association. We achieve this through single-view distillation, which stabilizes identity features within each camera, and cross-view reconstruction, which aligns feature spaces across cameras solely through visual consistency. Because the model relies only on RGB inputs, it remains robust to camera drift, reconfiguration, and diverse deployment conditions. Our contributions are summarized as follows:
\begin{enumerate}
    \item We introduce a fully self-supervised MCMOT framework that requires neither camera calibration nor manual annotations, enabling robust deployment in dynamic camera networks.
    \item We propose a distillation-and-reconstruction training scheme---combining single-view distillation and cross-view reconstruction---that improves cross-view identity consistency while retaining camera-dependent cues under significant viewpoint variations, without geometric constraints.
    \item We demonstrate strong generalization and state-of-the-art performance on two challenging benchmarks: on MMP-MvMHAT, our method improves overall accuracy by 3\% and F1 score by 7.5\%; on the more diverse MvMHAT, it achieves superior over-time identity stability and cross-view tracking performance.
\end{enumerate}
\section{Related Work}
\label{sec:related}

\subsection{Single-Camera Multi-Object Tracking}

Single-camera multi-object tracking (MOT) has been extensively studied, with the tracking-by-detection paradigm being the most widely adopted~\cite{cao2023observation, leal2016learning, schulter2017deep, wojke2017simple}. In this framework, object detectors~\cite{duan2019centernet, girshick2015fast, yolox2021} identify objects in each frame, and temporal associations are made using methods like the Kalman Filter~\cite{welch1995introduction} and the Hungarian Matching algorithm~\cite{kuhn1955hungarian}. Deep appearance features further improve association accuracy~\cite{chu2019famnet, xu2019spatial, xu2020train}. End-to-end approaches such as MOTR~\cite{zeng2022motr}, MOTRv2~\cite{zhang2023motrv2}, and TrackFormer~\cite{meinhardt2022trackformer} leverage query-based object detection to perform long-term tracking without manual association rules~\cite{carion2020end}. TransTrack~\cite{sun2020transtrack} and P3Aformer~\cite{zhao2022tracking} improve efficiency using location-based cost matrices. However, single-camera MOT struggles with occlusions and complex interactions due to limited viewpoints, motivating research in multi-camera multi-object tracking (MCMOT).

\subsection{Multi-Camera Multi-Object Tracking}

Multi-camera multi-object tracking (MCMOT) has gained growing attention for its complexity and broad applications. Existing methods fall into three main categories: distributed, global, and end-to-end. \textit{Distributed methods} perform tracking independently within each camera, followed by cross-view association~\cite{gilbert2006tracking, prosser2008multi, cai2014exploring, chen2014object}. Techniques such as hierarchical clustering~\cite{murtagh2012algorithms} and non-negative matrix factorization (NMF)~\cite{wang2012nonnegative} are used to merge intra-camera tracklets, though they often assume ideal conditions not met in dynamic environments. \textit{Global methods} detect individuals across views and associate them directly to build tracklets~\cite{ristani2016performance, maksai2017non, tesfaye2019multi}. TRACTA~\cite{he2020multi} and DMCT~\cite{you2020real} utilize perspective models and occupancy heatmaps, while ReST~\cite{cheng2023rest} employs a reconfigurable graph for robust association. \textit{End-to-end methods} like MUTR3D~\cite{zhang2022mutr3d}, PF-Track~\cite{pang2023standing}, and ViP3D~\cite{gu2023vip3d} are designed for 3D tracking tasks such as autonomous driving. MCTR~\cite{niculescu2024mctr} proposes a calibration-free framework using track embeddings, but it still relies on labeled data and assumes fixed camera setups, limiting flexibility in dynamic deployments. Our work focuses on overlapping-camera scenarios, where shared fields of view provide visual co-occurrence cues for cross-view association. In contrast, non-overlapping setups~\cite{javed2005appearance, tesfaye2017multi, chilgunde2004multi} face challenges such as time delays and lack of spatial correspondence.

\subsection{Self-Supervised Learning}

Self-supervised learning (SSL) has become a powerful paradigm in computer vision~\cite{doersch2015unsupervised, Noroozi2016UnsupervisedLO} and multimodal tasks~\cite{NEURIPS2021_cb3213ad, Wang2021MultimodalSL}, enabling robust representation learning without labeled data. Pretext tasks like context prediction~\cite{doersch2015unsupervised, pathak2016context}, jigsaw puzzles~\cite{Noroozi2016UnsupervisedLO, kim2018learning}, and colorization~\cite{larsson2017colorization} have shown effectiveness for image-level learning. For videos, pace prediction~\cite{10.1007/978-3-030-58520-4_30} and space-time cube puzzles~\cite{Kim2018SelfSupervisedVR} help capture temporal dynamics. More recent techniques such as contrastive learning~\cite{Chen2020ASF, He2019MomentumCF, Oord2018RepresentationLW} and masked autoencoding~\cite{He2021MaskedAA, Huang2022ContrastiveMA} are effective across both image and video domains. In multi-object tracking, self-supervised methods use spatial-temporal consistency to reinforce object identity. Strategies include cross-input consistency~\cite{NEURIPS2021_71e09b16}, cycle-consistency~\cite{Yin2023SelfsupervisedMT}, and path-consistency~\cite{Lu2024SelfSupervisedMT}. In MCMOT, recent approaches like MvMHAT~\cite{feng2024unveiling} and MvMHAT++~\cite{Gan2021SelfsupervisedMM} leverage consistency-based tasks such as symmetric-consistency (SymC) and transitive-consistency (TrsC), though they rely on CNN-based features. In contrast, our method adopts masked autoencoding~\cite{He2021MaskedAA}, which is more compatible with transformer architectures and enables richer, more adaptable representation learning in complex MCMOT settings.

\begin{figure*}[!tp]
    \centering
    \includegraphics[width=0.85\textwidth]{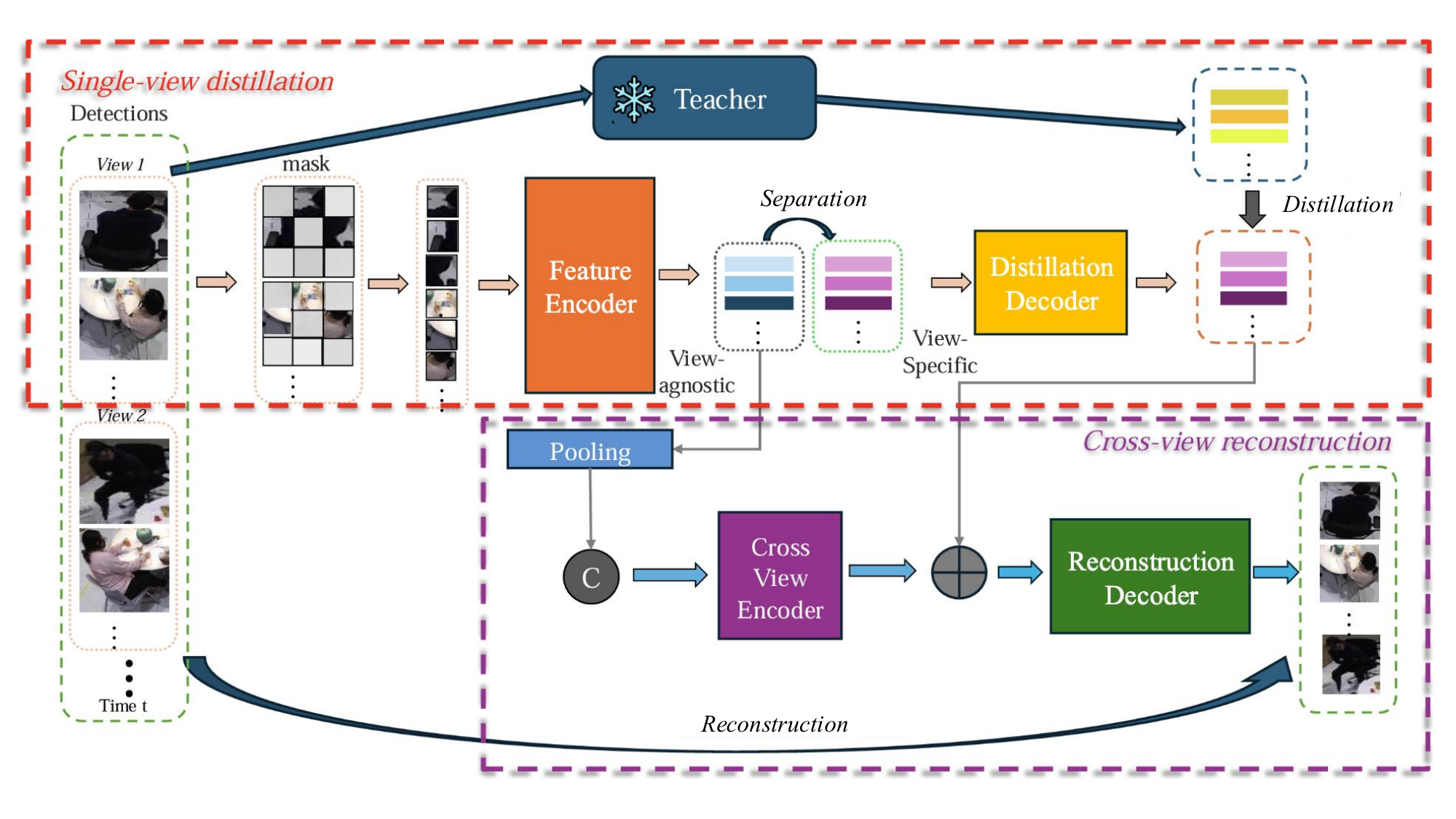}
    \vspace{-15pt}
    \caption{\textbf{Overview of CalibFree.} The method includes single-view distillation, feature separation, and cross-view reconstruction. In single-view distillation (red box), masked detections are encoded, and feature quality is supervised by a teacher model using a distillation loss. A separation regularizer encourages the learned features to specialize into view-agnostic and view-specific components. For cross-view reconstruction (purple box), pooled view-agnostic features are processed to reconstruct masked patches across views, optimized with a reconstruction loss.}
    \label{fig:pipeline} 
    \vspace{-15pt}
\end{figure*}
\section{Method}
In this section, we present the details of our proposed approach, \textbf{CalibFree}. We begin by formulating the problem, followed by a description of our algorithm, and conclude with how the generated features are used during inference.

\subsection{Problem Formulation}
\label{subsec:problem_formation}

Multi-camera multi-object tracking (MCMOT) aims to track all subjects across synchronized video streams from \( V \) cameras and associate identities across views. This can be formulated as a spatio-temporal association problem with two objectives:

\begin{itemize}
    \item \textbf{Intra-camera tracking:} Given detections \( D_t^v = \{D_i^v \mid i = 1, 2, \dots, N_t^v\} \) at frame \( t \) in view \( v \), associate them over time to form tracklets \( \tau_t^v \), as in single-camera MOT.
    \item \textbf{Cross-view matching:} Given the same detections at time \(t\), grouped by camera as \( \bar{D}_t = \{D_t^1, D_t^2, \dots, D_t^V\} \), match detections from different views that belong to the same subject.
\end{itemize}

Here, \(D_t^v\) and its appearance inside \( \bar{D}_t \) refer to the same underlying detections; the notation simply distinguishes their roles in temporal vs.\ cross-view association.

Like single-camera methods (e.g., DeepSORT~\cite{wojke2017simple}, ByteTrack~\cite{zhang2022bytetrack}), MCMOT relies on robust feature representations to ensure reliable association. These features must remain consistent across time and camera viewpoints while being discriminative enough to separate different identities.

Given all detections at time \( t \), \( D_t = \{D_{t,1}^1, \dots, D_{t,N_t^V}^V\} \), the goal is to extract two types of features for each detection \( D_{t,i}^v \):

\begin{itemize}
    \item \textbf{View-agnostic features} (\( f_a \)): Capture identity-preserving cues (e.g., silhouette, body shape, pose) for cross-view matching.
    \item \textbf{View-specific features} (\( f_s \)): Encode appearance-sensitive details (e.g., camera /style/illumination-dependent cues) useful for temporal tracking within a view.
\end{itemize}

These features support both within-view and cross-view association, enabling robust identity continuity across space and time in uncalibrated multi-camera environments.

\subsection{CalibFree}
\label{subsec:overview}

Masked autoencoders (MAEs)~\cite{He2021MaskedAA} motivate our design by showing that strong representations can be learned from partially observed inputs. CalibFree extends this idea to multi-camera tracking using a shared feature encoder that produces two complementary embeddings for each detection: a view-specific feature capturing fine-grained appearance cues, and a view-agnostic feature intended to remain consistent across cameras (Fig.~\ref{fig:pipeline}). The view-specific branch is refined using single-view distillation, while the view-agnostic branch is aligned across cameras through a cross-view reconstruction task that exploits overlapping fields of view. To encourage these two embeddings to play distinct roles, we introduce a redundancy-minimization regularizer during training that promotes minimal shared information between the two representations. We next detail the architecture, starting with the feature extraction module.

\textbf{Feature Extraction.}  
At each time step \(t\), synchronized frames from the \(V\) camera views are processed by a person detector to obtain bounding boxes. Each detected region is cropped, resized to \((H,W)\), and partitioned into \(M = (H/h)(W/w)\) non-overlapping patches. After patch embedding, each detection yields a sequence of \(M\) patch features in \(\mathbb{R}^E\).

A random mask with ratio \(\rho\) is applied independently to each detection, leaving \(M^{\text{vis}} = (1-\rho)M\) visible patches that are passed to the feature encoder. No patches from different detections or views are mixed at this stage. To maintain spatial correspondence for the later cross-view pathway, the same mask pattern (same masked patch indices) is applied across all camera views at time \(t\). Masked positions are represented by learned mask embeddings introduced only in decoder modules.

The visible patches for a single detection are fed into a ViT-based \emph{feature encoder}, which applies self-attention over these \(M^{\text{vis}}\) patches and outputs a set of patch-level features. These features are then split evenly into two components:
\[
f_a \in \mathbb{R}^{M^{\text{vis}} \times E/2}, \qquad 
f_s \in \mathbb{R}^{M^{\text{vis}} \times E/2}.
\]
The first component \(f_a\) serves as the view-agnostic representation, targeting identity-consistent cues that should hold across cameras. The second component \(f_s\) serves as the view-specific representation, capturing appearance details important for within-camera association.

To encourage these two representations to capture distinct aspects of the detection, we apply a separation regularizer during feature encoding. Specifically, a normalized mutual information (NMI) loss encourages \(f_a\) and \(f_s\) to share minimal information:
\[
L_{\text{sep}} = NMI(f_a, f_s).
\]
This regularizer promotes identity-consistent cues to be emphasized in the view-agnostic representation while preserving detailed appearance information in the view-specific representation, yielding complementary representations for the subsequent branches of CalibFree.

\textbf{Single-View Distillation.}  
The view-specific features \(f_s\) produced by the feature encoder capture fine-grained appearance cues---such as local texture and structure---that are crucial for reliable within-camera association but naturally vary across viewpoints. To strengthen these appearance-sensitive representations, we apply a single-view distillation process that transfers patch-level knowledge from a strong MAE-based teacher model pretrained on ImageNet.

For each detection, the encoder output \(f_s \in \mathbb{R}^{M^{\text{vis}} \times E/2}\) is first projected to the decoder dimension and combined with learned mask embeddings to construct a full length-\(M\) patch sequence aligned with the original layout. This sequence is passed through a lightweight ViT \emph{distillation decoder}, which predicts refined patch embeddings \(\hat{f}_s \in \mathbb{R}^{M \times E_d}\).

In parallel, the corresponding cropped detection image (with all \(M\) patches) is fed into a pretrained ViT-L MAE teacher to obtain patch-level teacher features. Since the MAE teacher has been pretrained extensively on large-scale single-view data (ImageNet), it provides stable and expressive appearance cues well suited for supervising within-view tracking. A linear projection maps the student outputs to the teacher feature space, and a Smooth L1 loss is applied:
\[
L_{\text{distill}} = \text{SmoothL1}(\hat{f}_s, f_{\text{teacher}}).
\]

This distillation pathway refines the view-specific representation to focus on detailed appearance information while leaving cross-camera consistent identity cues to be captured by the separate view-agnostic representation. As a result, \(f_s\) becomes a robust appearance descriptor for temporal association within each camera, without interfering with cross-view alignment or reconstruction.

\textbf{Cross-View Reconstruction.}  
The view-agnostic features \(f_a\) are intended to represent identity-consistent cues that should generalize across cameras. To promote such consistency, we introduce a cross-view reconstruction pathway that leverages complementary observations from multiple cameras at time \(t\) to reconstruct masked content in each detection.

For each detection, the patch-level view-agnostic features \(f_a \in \mathbb{R}^{M^{\text{vis}} \times E/2}\) are average-pooled to form a detection-level embedding \(g_a \in \mathbb{R}^{E/2}\). At time \(t\), we collect the embeddings from all detections across all \(V\) views and feed them into a shallow transformer \emph{cross-view encoder} that applies self-attention over the entire set. Crucially, this encoder does not rely on any identity correspondences; it treats all detections as candidates and learns relationships purely from the data. During training, the reconstruction objective encourages embeddings of detections that observe the same person to move closer together, because only those detections contain the complementary visual information needed to reconstruct each other’s masked regions. This produces refined, cross-view--aligned identity embeddings \(\hat{g}_a \in \mathbb{R}^{E_d}\) for every detection.

To reconstruct masked regions, the aligned embedding \(\hat{g}_a\) is broadcast to all patch locations and combined with the refined view-specific patch features \(\hat{f}_s\) from the single-view distillation pathway. These joint features form enriched per-patch representations containing both identity-level cues (shared across views) and appearance-level cues (specific to the current camera). A \emph{reconstruction decoder} uses these representations to predict pixel values for the masked patches of the original detection.

Because all cameras share the same patch mask pattern at time \(t\), the decoder can draw on information from unmasked regions in other views when reconstructing the masked regions of a given view. This forces the model to use the cross-view embedding \(\hat{g}_a\) to explain missing visual content across perspectives, thereby encouraging viewpoint-consistent identity representation without requiring geometry, calibration, or annotations.

The reconstruction loss is applied only to masked patches and uses mean squared error between the decoder output and the ground-truth pixels of the corresponding patches:
\[
L_{\text{recon}} = \text{MSE}\big(f^{\text{masked}}_{\text{reconstructed}}, 
f^{\text{masked}}_{\text{original}}\big).
\]
Through this cross-view reconstruction mechanism, CalibFree learns view-agnostic identity embeddings that are driven entirely by visual co-occurrence and complementary observations across cameras.

\textbf{Overall Training Objective.} CalibFree is optimized using three complementary losses introduced in the previous subsections:  
(1) a separation regularizer that encourages view-agnostic and view-specific representations to share minimal information,  
(2) a single-view distillation loss that refines appearance cues using a pretrained MAE teacher, and  
(3) a cross-view reconstruction loss that encourages cross-view identity consistency across cameras.

The overall training objective is the weighted sum of these components:
\[
L = L_{\text{sep}} + L_{\text{distill}} + L_{\text{recon}}.
\]

\subsection{Inference}
\label{subsec:inference}

During inference, only the feature encoder is used. Each detected person is cropped and passed through the encoder (without masking) to obtain a detection-level embedding, which is split into view-specific and view-agnostic components. The view-specific features drive within-camera tracking through a standard DeepSORT pipeline that begins new tracklets after short-term persistence, confirms them once stable, and removes them when inactive; motion is handled by Kalman filtering and appearance affinity by cosine similarity. Cross-camera association is performed once per time step by comparing the view-agnostic features of all active tracklets, constructing a pairwise similarity matrix, and applying bipartite matching to merge identities across cameras. When a cross-view match contradicts a local track ID, the global match takes precedence and the involved tracklets are unified to maintain consistent identity assignment. A global tracklet bank stores the most recent view-agnostic embedding for each global identity, enabling re-identification after long gaps or under sparse camera overlap. Together, this procedure yields coherent multi-camera tracking using only the learned encoder, without any geometric calibration, camera parameters, or identity annotations.
\section{Results}
Due to page limits, we include implementation details and more results in the Appendix.

\subsection{Datasets}
\begin{table*}[!t]
\centering
\resizebox{0.8\textwidth}{!}{
\begin{tabular}{c | c c c c c | c c c c | c c}
\toprule
 & \multicolumn{5}{c}{\textit{Over-time Tracking}} & \multicolumn{4}{c}{\textit{Cross-View Tracking}}  & \multicolumn{2}{c}{\textit{Overall}}\\ 
Methods & IDP & IDR & IDF1 & MOTA & HOTA & AIDP & AIDR & AIDF1 & MHAA & \textit{A} & \textit{F}\\
\midrule 
\multicolumn{12}{c}{\textit{supervised}}\\
Tracktor++\cite{Bergmann2019TrackingWB} & 67 & 56 & 61 & 67.2 & 46.2 & 62 & 23.2 & 33.8 & 19.1 & 43.1 & 47.4\\
CenterTrack\cite{Zhou2020TrackingOA} & 35.9 & 24.1 & 28.8 & 48.2 & 27.1 & 29.3 & 3.9 & 6.8 & 3.1 & 25.7 &17.8\\
TraDeS\cite{Wu2021TrackTD} & 59.7 & 50.1 & 54.5 & 66.1 & 42.7 & 54.5 & 17.3 & 26.2 & 13.3& 39.7 &40.4\\
TrackFormer\cite{meinhardt2022trackformer} & 41.8 & 28.6 & 34 & 46.7 & 30.2 & 39.9 & 5.3 & 9.4 & 3.2 & 25 & 21.7\\
DeepCC\cite{Ristani2018FeaturesFM} & 51.6 & 52.5 & 52.1 & 92.5 & 59.3 & 42.7 & 23.4 & 30.2 & 19.7 & 56.1 & 41.2\\
SVT\cite{Dong2021FastAR} & 63.1 & 63.4 & 63.3 & \underline{96.7} & 68.8 & 53.8 & 33.4 & 41.2 & 29.8 & 63.1 & 52.3\\

\midrule
\multicolumn{12}{c}{\textit{self-supervised}}\\
MvMHAT(YOLOX)\cite{Gan2021SelfsupervisedMM} & 51.1 & 53.2 & 52.7 & 82.1 & 47.2 & 30.4 & 17.1 & 23.2 & 14.1 & 48.1 & 37.9\\
MvMHAT(GT)\cite{Gan2021SelfsupervisedMM} & 58.6 & 58.8 & 58.7 & 93.7 & 65 & 35.4 & 21.2 & 26.5 & 20.3 & 57 & 42.6\\
MvMHAT++(YOLOX)\cite{feng2024unveiling} & 59.3 & 60.5 & 60.1 & 82.2 & 52.1 & 45.7 &34.1 & 40.5 & 28.9 & 55.5 & 50.3\\
MvMHAT++(GT)\cite{feng2024unveiling} & 67.1 & 67.6 & 67.3 & 95 &\underline{70.2} & \textbf{62.1} & 42.5 & \underline{50.4} & \underline{40.6} & \underline{67.8} &58.9\\

\midrule
\textbf{CalibFree(YOLOX)} & \underline{81.3} & \underline{77.1} & \underline{79.1} & 82.5 & 59.2 & 52.3 & \underline{48.4} & 50.3 & 34.4 & 58.4 & \underline{64.7}\\
\textbf{CalibFree(GT)} & \textbf{82.2} & \textbf{78} & \textbf{80} & \textbf{97.6}  & \textbf{75.7}  & \underline{55} & \textbf{50.9} & \textbf{52.8} & \textbf{44.2} & \textbf{70.8} & \textbf{66.4}\\

\bottomrule 
\end{tabular}
}
\caption{\textbf{Results on MMP-MvMHAT.} CalibFree surpasses both supervised and self-supervised methods across key metrics, demonstrating robust identity tracking in over-time and cross-view scenarios.}
\label{tab:results}
\vspace{-10pt}
\end{table*}
\begin{table*}[!t]
\centering
\resizebox{0.8\textwidth}{!}{
\begin{tabular}{c | c c c c c | c c c c | c c}
\toprule
 & \multicolumn{5}{c}{\textit{Over-time Tracking}} & \multicolumn{4}{c}{\textit{Cross-View Tracking}}  & \multicolumn{2}{c}{\textit{Overall}}\\ 
Methods & IDP & IDR & IDF1 & MOTA & HOTA & AIDP & AIDR & AIDF1 & MHAA & \textit{A} & \textit{F}\\
\midrule 
\multicolumn{12}{c}{\textit{supervised}}\\
Tracktor++\cite{Bergmann2019TrackingWB} & 54.2 & 40.1 & 46.1 & 66.5 & 42.8 & 34.3 & 14.6 & 20.5 & 37.1 & 51.8 & 33.3\\
CenterTrack\cite{Zhou2020TrackingOA} & 44.3 & 33.5 & 38.1 & 63.5 & 37.8 & 29.7 & 9.1 & 13.9 & 34.1 & 48.8 & 26.0\\
TraDeS\cite{Wu2021TrackTD} & 46.7 & 43.2 & 44.9 & \underline{69.5} & 42.9 & 32.4 & 14.0 & 19.6 & 36.0 & 52.8 & 32.2\\
TrackFormer\cite{meinhardt2022trackformer} & 52.3 & 47.2 & 49.6 & \textbf{70.4} & 47.3 & 47.8 & 23.2 & 31.3 & 40.2 & 55.3 & 40.4\\
DeepCC\cite{Ristani2018FeaturesFM} & 44.7 & 44.2 & 44.4 & 63.9 & 41.1 & 57.9 & 34.8 & 43.4 & 43.8 & 53.9 & 43.9\\
SVT\cite{Dong2021FastAR}  & 47.9 & 47.2 & 47.6 & 65.4 & 43.1 & \underline{61.7} & 45.7 & 52.5 & 50.4 & 56.9 & 50.0\\

\midrule
\multicolumn{12}{c}{\textit{self-supervised}}\\
MvMHAT\cite{Gan2021SelfsupervisedMM}& 53.1 & 52.0 & 52.5 & 64.7 & 47.9 & 53.0 & 46.4 & 49.5 & 51.7 & 58.2 & 51.0\\
MvMHAT++\cite{feng2024unveiling} & \underline{58.5} & \underline{57.4} & \underline{57.9} & 66.3 & 51.8 & \textbf{63.8} & \underline{56.0} & \textbf{59.6} & \textbf{59.7} & \textbf{63.0} & \textbf{58.8}\\

\midrule
\textbf{CalibFree} & \textbf{59.1} & \textbf{58.4} & \textbf{58.7} & 60.4  & \textbf{52.0}  & 58.9 & \textbf{56.2} & \underline{57.4} & \underline{57.0} & \underline{58.4} & \underline{58.1}\\

\bottomrule 
\end{tabular}
}

\caption{\textbf{Results on MvMHAT.} CalibFree achieves best or second best results across most key metrics.}
\label{tab:mvmhat}
\vspace{-10pt}
\end{table*}

\noindent\textbf{MMP-MvMHAT.}
Adapted from MMPTRACK~\cite{mmptrackchallenge}, MMP-MvMHAT features 4--6 overlapping indoor cameras and 28 individuals. It provides 8{,}000 fully annotated frames across four training scenes and 4{,}000 frames for validation, with no calibration data. This setup, focused on crowded and occluded environments, poses a challenging multi-view tracking task.

\noindent\textbf{MvMHAT.}
MvMHAT~\cite{feng2024unveiling} is a large-scale dataset containing 26 video groups (98 sequences) sourced from Campus~\cite{Xu_2016_CVPR}, EPFL~\cite{4359319}, and newly collected footage. Each group includes 3--4 synchronized camera views, totaling over 90{,}000 annotated frames. Split into training and testing (13 groups each) with a 2:1 ratio, MvMHAT covers diverse scenarios and camera angles---often with 90$^\circ$ viewpoint differences---to facilitate robust multi-view tracking evaluation.

\subsection{Evaluation Metrics}
\noindent\textbf{Over-time Tracking.}
We adopt Multiple Object Tracking Accuracy (MOTA)~\cite{mota} to assess single-view tracking performance in terms of false positives, missed detections, and identity switches. Given the emphasis on robust identity association over time, we further use ID Precision (IDP), ID Recall (IDR), and ID F1 (IDF1)~\cite{ristani2016performance}, as well as High Order Tracking Accuracy (HOTA)~\cite{hota} for a balanced evaluation of detection, association, and localization.

\noindent\textbf{Cross-view Tracking.}
For multi-camera scenarios, we use Association ID Precision (AIDP), Association ID Recall (AIDR), and Association ID F1 (AIDF1)~\cite{AIDF1_2020,AIDF1_2021}, which average pairwise matching accuracy across different cameras. We also include Multi-view Multi-Human Association Accuracy (MHAA), which penalizes identity-consistency errors in multi-camera contexts with frequent occlusions and appearance shifts.

\noindent\textbf{Overall.}
To provide a holistic MCMOT assessment, we calculate the MCMOT F1 score ($\textit{F}$) and accuracy score ($\textit{A}$) by taking the average of F1 and accuracy across both over-time and cross-view tracking~\cite{feng2024unveiling}:
\[
\textit{F} = \text{Mean(IDF1, AIDF1)} \quad,\quad \textit{A} = \text{Mean(MOTA, MHAA)}.
\]

\subsection{Main Results and Analysis}

\subsubsection{Baseline Methods}

We compare our method against state-of-the-art approaches in Tables~\ref{tab:results} and~\ref{tab:mvmhat}, where the best results are highlighted and second-best underlined.

For single-camera (over-time) tracking, we include four representative MOT methods: Tracktor++~\cite{Bergmann2019TrackingWB}, CenterTrack~\cite{Zhou2020TrackingOA}, TraDeS~\cite{Wu2021TrackTD}, and TrackFormer~\cite{meinhardt2022trackformer}. Since these do not support cross-view tracking, we assign ground-truth IDs upon first appearance in each camera and apply each method independently within views.

For MCMOT, we evaluate DeepCC~\cite{Ristani2018FeaturesFM} and SVT~\cite{Dong2021FastAR}, with DeepCC leveraging an off-the-shelf ReID model~\cite{DBLP:journals/corr/abs-1711-10295} for cross-view association. We also include MvMHAT and its extension MvMHAT++, two self-supervised methods that require no fine-tuning, with MvMHAT++ introducing an additional training stage. All MCMOT methods are evaluated on the MMP-MvMHAT dataset using ground-truth bounding boxes for consistency, except that previous self-supervised methods are also tested with YOLOX-generated detections.

Detector-based results using YOLOv7 and YOLOX are shown in Table~\ref{tab:detector}. For MvMHAT, we use a Detectron2~\cite{wu2019detectron2} detector; among available models, we select the ResNet-50 variant that achieves MOTA closest to the original MvMHAT paper for fair comparison.

Lastly, we do not re-train supervised baselines on our dataset, as they require labeled data---unlike our self-supervised approach---ensuring a fair comparison in terms of generalization and calibration-free capability.

\begin{table*}[!t]
\centering
\resizebox{0.8\textwidth}{!}{
\begin{tabular}{c | c c c c c | c c c c | c c}
\toprule
 & \multicolumn{5}{c}{\textit{Over-time Tracking}} & \multicolumn{4}{c}{\textit{Cross-View Tracking}} & \multicolumn{2}{c}{\textit{Overall}} \\ 
Methods & IDP & IDR & IDF1 & MOTA & HOTA & AIDP & AIDR & AIDF1 & MHAA & \textit{A} & \textit{F}\\
\midrule 

ViT-L (teacher) & 82.1 & 77.8 & 79.9 & 97.5 & 75.6 & 47.9 & 44.4 & 46.1 & 36.7 & 67.1 & 63\\
Distillation only & 78.4 & 67.8 & 72.7 & 97.2 & 68.4 & 47.5 & 40.1 & 43.5 & 34.4 & 65.8 & 58.1\\
Reconstruction only& 81.2 & 77.1 & 79.1 & 97.5 & 75.2 & 52.3 & 48.4 & 50.3 & 41.8 & 69.7 & 64.7\\

\midrule
\textbf{CalibFree(ours)} & \textbf{82.2} & \textbf{78} & \textbf{80} & \textbf{97.6}  & \textbf{75.7}  & \textbf{55} & \textbf{50.9} & \textbf{52.8} & \textbf{44.2} & \textbf{70.8} & \textbf{66.4}\\

\bottomrule 
\end{tabular}
}

\caption{\textbf{Ablation studies of CalibFree variations.} The full model, combining distillation, reconstruction, and feature separation, achieves the best performance across all tracking metrics.}
\label{tab:effectiveness}
\vspace{-10pt}
\end{table*}
\begin{table*}[!t]
\centering
\resizebox{0.8\textwidth}{!}{
\begin{tabular}{c c | c c c c c | c c c c | c c}
\toprule
 Detector & Detector & \multicolumn{5}{c}{\textit{Over-time Tracking}} & \multicolumn{4}{c}{\textit{Cross-View Tracking}} & \multicolumn{2}{c}{\textit{Overall}}\\ 
Pretrain & Inference & IDP & IDR & IDF1 & MOTA & HOTA & AIDP & AIDR & AIDF1 & MHAA & \textit{A} & \textit{F}\\
\midrule 
YOLO7 & YOLO7 & 59.9 & 57.5 & 58.6 & 44.9 & 47.7 & 47.2 & 44.3 & 45.7 & 16.4 & 30.7 & 52.2\\
Groundtruth & YOLO7 & 59.8 & 57.4 & 58.5 & 44.9 & 47.7 & 49.3 & 46.1 & 47.6 & 18.3 & 31.6 & 53.1\\
YOLOX & YOLOX & 81.3 & 77.1 & 79.1 & 82.5 & 59.2 & 52.3 & 48.4 & 50.3 & 34.4 & 58.4 & 64.7\\
Groundtruth & YOLOX & 81 & 76.8 & 78.9 & 82.5 & 59 & 54 & 50.1 & 52 & 37.5 & 60 & 65.5\\

\textbf{Groundtruth} & \textbf{Groundtruth} & \textbf{82.2} & \textbf{78} & \textbf{80} & \textbf{97.6}  & \textbf{75.7}  & \textbf{55} & \textbf{50.9} & \textbf{52.8} & \textbf{44.2} & \textbf{70.8} & \textbf{66.4}\\

\bottomrule 
\end{tabular}
}

\caption{\textbf{Ablation study on detector choice during pretraining and inference.} Results show that CalibFree maintains high ID association consistency, with inference detector choice impacting tracking accuracy more than pretraining.}
\vspace{-10pt}
\label{tab:detector}
\end{table*}

\subsubsection{Results on MMP-MvMHAT}
\begin{itemize}
    \item \textbf{Over-Time Tracking:} Table~\ref{tab:results} reports CalibFree's performance on the indoor-focused MMP-MvMHAT dataset, where many subjects exhibit limited motion (e.g., seated office workers). Despite the simplicity of such motion---often inflating IDF1 for other methods---CalibFree achieves an IDF1 of 80.0, demonstrating strong identity continuity under occlusion and crowding. Its HOTA score of 75.7 reflects balanced accuracy in detection and association, minimizing ID switches and ensuring stable long-term tracking.

    \item \textbf{Cross-View Tracking:} CalibFree achieves an AIDF1 of 52.8 and MHAA of 44.2, outperforming all baselines. While its AIDP is slightly lower than MvMHAT++, CalibFree achieves higher AIDR, indicating better recall of cross-camera matches. This reflects its ability to capture identity-consistent cues under challenging viewpoint changes and appearance similarity. Compared to supervised baselines like DeepCC and SVT---which require manual annotations---CalibFree delivers stronger association without labels. Center-based trackers (e.g., CenterTrack) lack robust appearance modeling and accumulate ID switches in crowded scenes, while TrackFormer can produce mismatches when its detection step underperforms.

    \item \textbf{Sensitivity to Bounding Box Quality:} As shown in Table~\ref{tab:results}, CalibFree is more robust to noisy bounding boxes than prior self-supervised methods using YOLOX detections. While others suffer significant performance drops, CalibFree maintains ID-related metrics with minimal degradation, highlighting the resilience of its learned features to imperfect detections.

    \item \textbf{Overall:} CalibFree surpasses self-supervised baselines (MvMHAT, MvMHAT++) in both accuracy (70.8) and F1 score (66.4), demonstrating the advantage of its specialized view-agnostic and view-specific representations for both temporal and cross-view association. Although motion in MMP-MvMHAT is simpler, the cluttered indoor layout introduces frequent identity ambiguities, which CalibFree handles effectively.
\end{itemize}

\subsubsection{Results on MvMHAT}
\begin{itemize}
    \item \textbf{Over-Time Tracking.} Table~\ref{tab:mvmhat} presents results on the MvMHAT dataset, which includes both indoor and outdoor scenes with sparsely placed cameras and minimal overlap. Single-camera methods like TrackFormer yield competitive MOTA (e.g., 70.4), largely reflecting detector quality. However, CalibFree achieves higher ID-based metrics (IDP, IDR, IDF1) and HOTA, indicating better temporal identity consistency.
    
    \item \textbf{Cross-View Tracking.} Single-view trackers struggle when ID switches occur, propagating errors across views and degrading AIDF1. That said, their cross-view performance improves slightly over MMP-MvMHAT due to reduced overlap and fewer direct transitions. CalibFree outperforms multi-view trackers like DeepCC and SVT across all cross-view metrics. MvMHAT++ achieves higher precision and AIDF1 in this setting, which we attribute to two factors: (1) CalibFree uses a less accurate detector (lower MOTA), reducing cross-view consistency; and (2) MvMHAT++ benefits from a second-stage training step specifically tailored for cross-view association, offering an advantage in sparsely overlapped environments.
    
    \item \textbf{Overall Performance.} CalibFree demonstrates strong gains over most single-camera and multi-view baselines in overall accuracy ($A$) and F1 score ($F$). By learning specialized view-agnostic and view-specific representations, it maintains identity across views without calibration. Although MvMHAT++ excels in some cross-view metrics, CalibFree's unified, annotation-free framework delivers robust and generalizable performance under real-world challenges like occlusions, sparse views, and subject similarity.
\end{itemize}

\begin{table}[t]
\centering
\small
\resizebox{0.8\textwidth}{!}{%
\begin{tabular}{lccc}
\toprule
Feature & Camera ID Acc. (\%) $\uparrow$ & Intra-Cam IDF1 $\uparrow$ & Cross-Cam IDF1 $\uparrow$ \\
\midrule
$f_a$ (view-agnostic)     & 33.6 & 71.8 & 64.2 \\
$f_s$ (view-specific)     & 78.4 & 74.9 & 52.6 \\
Merged ($f_a + f_s$)      & 61.3 & 73.8 & 58.6 \\
$f_a \oplus f_s$          & N/A  & 74.2 & 59.7 \\
\bottomrule
\end{tabular}}
\caption{\textbf{Camera-sensitivity probe.} $f_s$ is more camera-dependent and favors intra-camera association, while $f_a$ suppresses camera cues and favors cross-camera matching. \textit{Merged} uses a single embedding (no feature separation); $f_a \oplus f_s$ routes $f_a$/$f_s$ to cross-/intra-camera association (probing not applicable).}
\label{tab:probing}
\vspace{-6pt}
\end{table}
\begin{table*}[!t]
\centering
\resizebox{0.8\textwidth}{!}{
\begin{tabular}{c c | c c c c c | c c c c | c c}
\toprule
Teacher & Student & \multicolumn{5}{c}{\textit{Over-time Tracking}} & \multicolumn{4}{c}{\textit{Cross-View Tracking}} & \multicolumn{2}{c}{\textit{Overall}} \\ 
Model & Model & IDP & IDR & IDF1 & MOTA & HOTA & AIDP & AIDR & AIDF1 & MHAA & \textit{A} & \textit{F}\\
\midrule 

\textbf{ViT-L} & \textbf{ViT-B} &\textbf{82.2} & \textbf{78} & \textbf{80} & \textbf{97.6}  & \textbf{75.7}  & \textbf{55} & \textbf{50.9} & \textbf{52.8} & \textbf{44.2} & \textbf{70.8} & \textbf{66.4}\\
ViT-B & ViT-B & 82.0 & 77.8 & 79.8 & 97.5 & 75.5 & 52.1 & 47.8 & 49.7 & 42.0 & 69.8 & 64.8\\
ViT-B & ViT-S & 77.6 & 75.8 & 76.7 & 97.2 & 71.8 & 50.3 & 45.4 & 47.7 & 40.4 & 68.8 & 62.2\\

\bottomrule 
\end{tabular}
}

\caption{\textbf{Ablation studies of different models.} Using ViT-L as teacher and ViT-B as student achieves best results.}
\label{tab:teacher}
\vspace{-10pt}
\end{table*}
\begin{table*}[!t]
\centering
\resizebox{0.8\textwidth}{!}{
\begin{tabular}{c | c c c c c | c c c c | c c}
\toprule
 & \multicolumn{5}{c}{\textit{Over-time Tracking}} & \multicolumn{4}{c}{\textit{Cross-View Tracking}} & \multicolumn{2}{c}{\textit{Overall}} \\ 
Mask ratio & IDP & IDR & IDF1 & MOTA & HOTA & AIDP & AIDR & AIDF1 & MHAA & \textit{A} & \textit{F}\\
\midrule 
0.9 & \textbf{82.3} & 78 & \textbf{80.1} & 97.6  & \textbf{75.8}  & 53.8 & 48.7 & 51.1 & 43.0 & 70.3 & 65.6 \\
\textbf{0.75} & 82.2 & \textbf{78} & 80 & \textbf{97.6}  & 75.7  & 55 & \textbf{50.9} & \textbf{52.8} & \textbf{44.2} & \textbf{70.8} & \textbf{66.4}\\
0.5 & 82.0 & 77.9 & 79.8 & 97.5  & 75.6  & \textbf{55.1} & 50.1 & 52.4 & 43.9 & 70.7 & 66.1 \\

\bottomrule 
\end{tabular}
}

\caption{\textbf{Ablation studies of mask ratios.} 0.75 achieves the best balance between over-time and cross-view tracking.}
\label{tab:mask_ratio}
\vspace{-10pt}
\end{table*}

\subsection{Ablation Studies}
\subsubsection{Effect of Distillation and Reconstruction}

Table~\ref{tab:effectiveness} presents an ablation study comparing four CalibFree variants. The \textit{ViT-L (teacher)} setting uses features from a pretrained ViT-L~\cite{dosovitskiy2020vit} directly, without training a student. \textit{Distillation only} trains a student using only the distillation loss \(\mathcal{L}_{\text{distill}}\), while \textit{Reconstruction only} trains a student from scratch using only \(\mathcal{L}_{\text{recon}}\). Our \textit{full model} combines \(\mathcal{L}_{\text{distill}} + \mathcal{L}_{\text{recon}} + \mathcal{L}_{\text{sep}}\) and learns specialized view-specific and view-agnostic features.

The ViT-L teacher achieves strong over-time tracking (IDF1: 79.9, MOTA: 97.5) but limited cross-view performance (AIDF1: 46.1, MHAA: 36.7). Distillation alone underperforms due to reduced model capacity and masked inputs (AIDF1: 43.5). Reconstruction alone slightly lowers over-time performance (IDF1: 79.1) but improves cross-view accuracy (AIDF1: 50.3), highlighting the importance of spatial reconstruction for multi-view consistency. The full CalibFree model achieves the best overall results (F: 66.4, Accuracy: 70.8), validating the importance of all components for robust uncalibrated tracking.

\subsubsection{Effect of Detector Choice}

Table~\ref{tab:detector} compares three detector configurations---ground truth, YOLOX~\cite{yolox2021}, and YOLOv7~\cite{wang2023yolov7}---used during both pretraining and inference. CalibFree demonstrates strong robustness to detector choice during pretraining; however, inference quality has a more substantial effect. Switching from YOLOX to the less accurate YOLOv7 results in noticeable performance drops, highlighting the importance of reliable detections---a challenge common to all tracking methods. Notably, models pretrained with ground truth and inferred using YOLOX achieve ID-based metrics comparable to those with ground-truth inference, demonstrating CalibFree's adaptability when the inference detector maintains reasonable accuracy. Moreover, as shown in Table~\ref{tab:results}, CalibFree is significantly more resilient to bounding box imperfections compared to previous self-supervised methods.

\subsubsection{Camera-Sensitivity Probe.}
We provide quantitative evidence of feature separation using a camera-sensitivity probe. Specifically, we train a linear classifier to predict camera ID from frozen embeddings on 1{,}000 frames from MMP-MvMHAT scenes with four cameras. As shown in Table~\ref{tab:probing}, the view-specific feature $f_s$ achieves substantially higher camera-ID accuracy, whereas the view-agnostic feature $f_a$ remains only slightly above chance (25\% for 4 cameras). This indicates that camera-dependent information is primarily captured by $f_s$, while $f_a$ suppresses camera cues and better supports cross-camera association.

\subsubsection{Impact of Teacher and Student Model Sizes}
Table~\ref{tab:teacher} investigates three teacher--student setups: ViT-L$\to$ViT-B, ViT-B$\to$ViT-B, and ViT-B$\to$ViT-S. A larger teacher (ViT-L) improves cross-view performance (AIDF1: 49.7$\to$52.8, MHAA: 42.0$\to$44.2) due to richer representations feeding into the cross-view encoder. Over-time metrics (IDF1, MOTA) remain mostly unchanged, suggesting that temporal continuity is less sensitive to teacher size. On the other hand, reducing the student to ViT-S lowers both over-time (IDF1: 76.7) and cross-view (AIDF1: 47.7) performance, indicating insufficient capacity for robust identity modeling. Although smaller students are more efficient, this comes at the cost of accuracy---especially in complex multi-view scenarios.

\subsubsection{Effect of Mask Ratios in Pretraining}
We examine the impact of mask ratios (0.5, 0.75, 0.9) in Table~\ref{tab:mask_ratio}. High masking (e.g., 0.9) harms cross-view performance (AIDF1: 51.1, MHAA: 43.0), suggesting that excessive masking limits the model's ability to learn spatially consistent features. Over-time metrics (IDF1, MOTA) remain stable, as temporal tracking depends less on detailed spatial information. While both 0.5 and 0.75 achieve comparable cross-view results (AIDF1: 52.4 vs. 52.8), the 0.75 setting reduces token usage, offering better efficiency. Thus, a 0.75 mask ratio strikes the best balance between accuracy and computational cost for both temporal and cross-view tracking.
\section{Conclusion, Limitation, and Future Work}
We have introduced \textbf{CalibFree}, a self-supervised multi-camera multi-object tracking (MCMOT) method that achieves state-of-the-art performance without relying on camera calibration or manual annotations. By encouraging a separation of view-agnostic and view-specific features---supported by cross-view reconstruction and knowledge distillation---CalibFree robustly handles complex identity association across time and views. Experiments on the MMP-MvMHAT and MvMHAT datasets underscore its strong adaptability in both over-time and cross-view tracking. Despite these advances, our approach remains limited by its exclusive use of RGB signals, thereby overlooking potentially useful geometric relationships across views. Although learning geometry without camera parameters is nontrivial, incorporating geometric cues in a self-supervised manner could further enhance identity consistency and cross-view association. In future work, we aim to explore self-supervised strategies for integrating such geometric cues, enabling CalibFree to better leverage spatial relationships between detections under dynamic camera configurations.

\clearpage
\clearpage
\setcounter{page}{1}
\setcounter{section}{0}

\begin{figure}[h]
    \centering
    \includegraphics[width=\columnwidth]{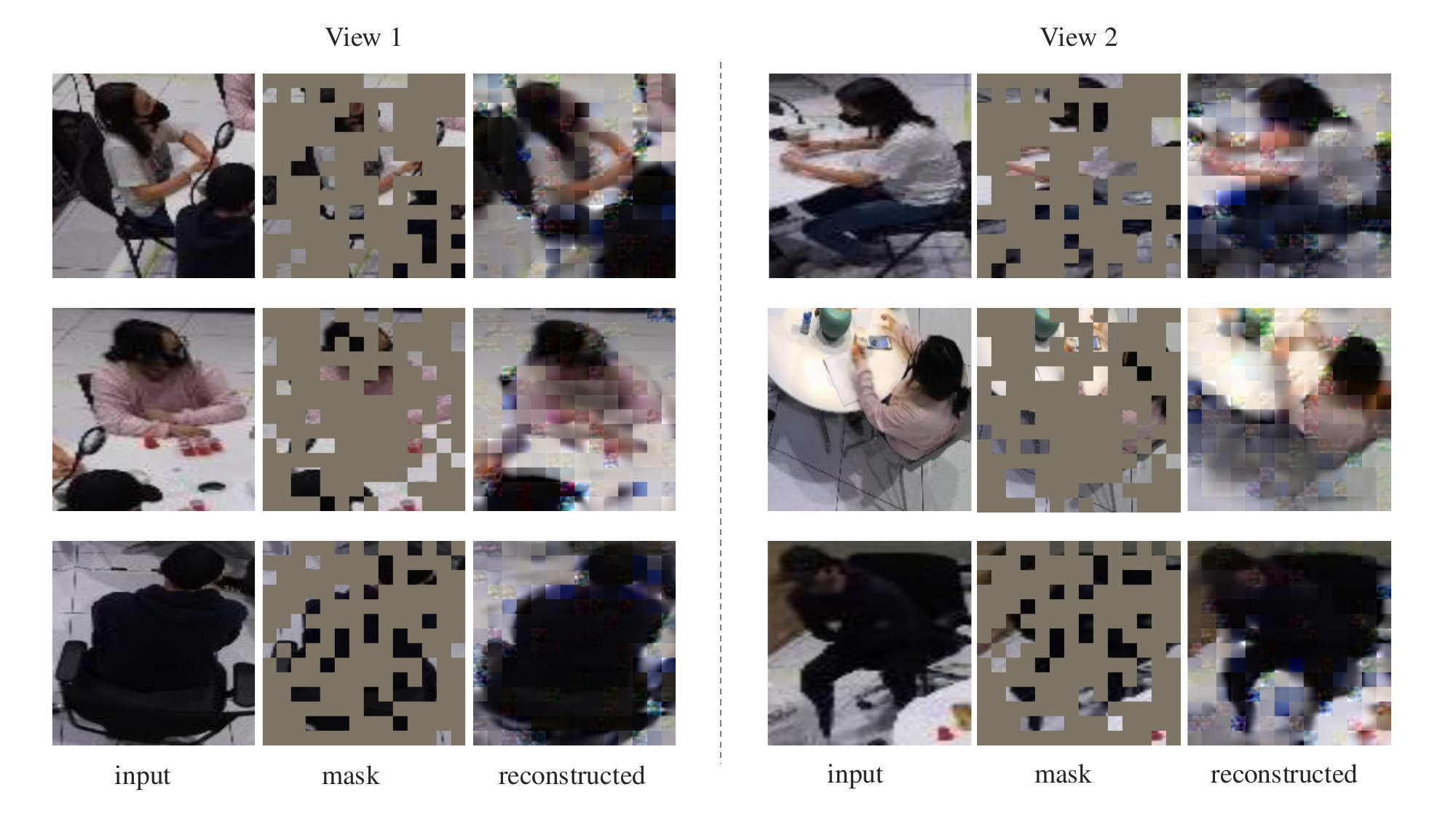}
    \vspace{-15pt}
    \caption{\textbf{Cross-view reconstruction results.} The input images show the original crops, while the masked images indicate regions removed for reconstruction. CalibFree reconstructs the masked regions by leveraging complementary observations across views, recovering consistent appearance and identity-relevant cues even when large portions are obscured.}
    \label{fig:visual}
    \vspace{-12pt}
\end{figure}

\noindent\textbf{Calibration independence.} CalibFree never accesses camera intrinsics, extrinsics, or homographies during training or inference, and thus requires no calibration when deployed to new camera networks.

\section{Implementation Details}  

\textbf{Pretraining Phase.}  
After detecting the bounding box for each person, the region of interest (ROI) is cropped based on the bounding box coordinates and resized to a fixed resolution of \(224 \times 224\) pixels through upsampling or downsampling. These resized ROIs are divided into non-overlapping patches of size \(16 \times 16\).  

The \textit{feature encoder} is implemented as a vanilla Vision Transformer (ViT) base model with 12 transformer blocks and 12 attention heads, using an embedding dimension of 768. During inference, we aggregate per-patch features (average pooling) to form a detection-level embedding, which is then split into view-agnostic and view-specific components (each 384-d for ViT-B).  

The \textit{distillation decoder}, \textit{cross-view encoder}, and \textit{reconstruction decoder} are shallow Vision Transformer models, each comprising 8 attention blocks and 16 attention heads with an embedding dimension of 512. For knowledge distillation, the \textit{teacher model} is a ViT large model pretrained on ImageNet, with an output feature dimension of 1024. To align dimensions during distillation, the 512-dimensional output of the distillation decoder is projected to 1024 dimensions using a linear layer.  

Pretraining is conducted over 400 epochs with a base learning rate of \(1.5 \times 10^{-4}\) and a 40-epoch warmup phase. The weight decay is set to 0.05. Reconstruction loss is computed using mean squared error (MSE) loss applied to normalized pixel values rather than raw pixel values. The maximum number of detections is set to be 10 for both datasets. 

\begin{table*}[t]
\centering
\resizebox{0.88\textwidth}{!}{
\begin{tabular}{c c | c c c c c | c c c c | c c}
\toprule
 & \multicolumn{5}{c}{\textit{Over-time Tracking}} & 
   \multicolumn{4}{c}{\textit{Cross-View Tracking}}  & 
   \multicolumn{2}{c}{\textit{Overall}}\\ 
Camera & Effect & IDP & IDR & IDF1 & MOTA & HOTA & AIDP & AIDR & AIDF1 & MHAA & \textit{A} & \textit{F}\\
\midrule 

2 & Misalignment 
  & 81.9 & 78.0 & 79.8 & 97.3 & 74.9 
  & 54.8 & 50.7 & 52.7 & 43.9 
  & 70.4 & 66.3 \\

2 & Malfunction 
  & 77.6 & 66.6 & 71.7 & 95.0 & 72.8 
  & 50.4 & 46.9 & 48.6 & 42.0 
  & 68.5 & 60.2 \\
\midrule 

1 & Misalignment 
  & 82.1 & 78.0 & 80.0 & 97.4 & 75.3 
  & 55.1 & 50.7 & 52.8 & 44.0 
  & 70.7 & 66.4 \\

1 & Malfunction 
  & 79.4 & 74.5 & 76.9 & 96.2 & 73.9 
  & 53.4 & 48.9 & 51.1 & 43.0 
  & 69.6 & 64.0 \\
\midrule

0 & N/A  
  & \textbf{82.2} & \textbf{78.0} & \textbf{80.0} & \textbf{97.6} & \textbf{75.7}
  & \textbf{55.0} & \textbf{50.9} & \textbf{52.9} & \textbf{44.2} 
  & \textbf{70.9} & \textbf{66.5} \\

\bottomrule 
\end{tabular}
}

\caption{\textbf{Results on MMP-MvMHAT with dynamic camera configurations.}}
\label{tab:dynamic}
\vspace{-10pt}
\end{table*}

\noindent\textbf{Inference Phase.}  
During inference, a detector identifies bounding boxes for all persons, and the ROIs are cropped based on the bounding box coordinates. Features for each detection are generated by feeding all patches (without masking) to the feature encoder. The per-patch features are then aggregated using average pooling to produce a detection-level embedding, which is split into view-agnostic and view-specific components.  

For \textit{over-time tracking}, we utilize DeepSORT, with the generated features serving as the primary matching criterion, complemented by a Kalman filter as a secondary criterion. For \textit{cross-view tracking}, we compute pairwise similarity across views and apply bipartite matching to merge identities (see Section~\ref{subsec:inference}).  

\section{Visualization}
Figure~\ref{fig:visual} illustrates how CalibFree captures identity-preserving features across different viewpoints, showing the original input, masked patches, and reconstructed outputs. Substantial portions of each subject are masked to simulate partial observations, yet CalibFree reliably restores these regions by leveraging both view-agnostic and view-specific features. Crucial details like posture, clothing texture, and overall silhouette remain largely intact, supporting consistent identity tracking across camera views. Even when significant information is obscured, the model reconstructs occluded areas accurately based on the available visible patches, all without requiring camera calibration. This visualization underscores CalibFree's robustness and practical utility in multi-camera scenarios with frequent occlusions and substantial viewpoint variations. 

\section{Computation and Runtime}
\label{sec:compute_runtime}

We report the hardware setup and training time for each dataset:

\begin{table}[h]
\centering
\resizebox{0.85\columnwidth}{!}{%
\begin{tabular}{lcccccc}
\toprule
\textbf{Dataset} & \textbf{\#\,GPUs} & \textbf{GPU Model} & \textbf{Mem/GPU} & \textbf{Batch} & \textbf{Epochs} & \textbf{Wall-clock} \\
\midrule
MMP-MvMHAT & 4 & NVIDIA H100 & 80\,GB & 5 & 400 & 13.4 h \\
MvMHAT     & 4 & NVIDIA A100 & 80\,GB & 5 & 400 & 38.3 h \\
\bottomrule
\end{tabular}}
\end{table}

At inference, using four cameras with 1080P input resolution, a ViT-B encoder (as feature encoder) for each view, and DeepSORT as the tracker, the average runtime per timestep is 154 ms ($\pm$ 32 ms) on one NVIDIA A100 80GB GPU.

\section{Dynamic Camera Setting}

As discussed in Section~\ref{sec:intro} and visualized in Figure.~\ref{fig:vis_data}, long-term real-world deployments often cause cameras to become gradually misaligned or even partially fail. To assess the robustness of our method under such conditions, we design controlled perturbation experiments on the MMP-MvMHAT dataset. Since the dataset does not provide ground-truth camera calibration or configuration metadata, we simulate \textit{camera misalignment} by randomly cropping 90\% of a camera's input, realigning the bounding boxes, and interpolating the view back to its original resolution. This reduces the effective field of view and mimics the loss of scene coverage caused by physical camera displacement. To simulate \textit{camera malfunction}, we randomly select camera IDs and mask all frames originating from those cameras, representing complete sensor failure. To isolate the effect of viewpoint corruption rather than detection errors, we evaluate all experiments using ground-truth bounding boxes.

Table~\ref{tab:dynamic} reports tracking performance under these dynamic camera configurations, with the last row corresponding to the original CalibFree results without perturbation. We examine scenarios where one or two cameras experience misalignment or malfunction. As shown, our method remains highly robust to camera misalignment, consistent with the design of CalibFree, which relies solely on RGB appearance without requiring camera parameters. The method also demonstrates strong resilience to partial camera malfunction. Although performance drops more noticeably when two cameras fail, we attribute this degradation primarily to the reduced amount of multi-view training data, as self-supervised association learning still requires sufficiently diverse cross-view observations to extract reliable representations. Nevertheless, this limitation can be mitigated in practice by collecting additional temporal data from the remaining operational cameras, which gradually compensates for the reduced viewpoint diversity.

\clearpage
\bibliographystyle{splncs04}
\bibliography{main}

\end{document}